\documentclass[BCOR=1mm, DIV=calc,10pt,
twoside=true,
twocolumn,
headings=normal]{scrartcl}

\usepackage{microtype}
\usepackage{graphicx}
\usepackage{subfigure}
\usepackage{booktabs} 

\usepackage{amsmath}
\usepackage{bm}
\usepackage{url}

\usepackage{cite}
\usepackage{amsmath,amssymb,amsfonts}
\usepackage{algorithmic}
\usepackage{graphicx}
\usepackage{textcomp}
\usepackage{xcolor}

\usepackage{authblk}
\begin{document}

\title{Cyclic Boosting - an explainable supervised machine learning algorithm}

\author[1]{Felix Wick \thanks{felix.wick@blueyonder.com}}
\author[2]{Ulrich Kerzel \thanks{u.kerzel@iubh-fernstudium.de}}
\author[1]{Michael Feindt \thanks{michael.feindt@blueyonder.com}}
\affil[1]{\small Blue Yonder GmbH (Karlsruhe, Germany)}
\affil[2]{\small IUBH Internationale Hochschule (Erfurt, Germany)}


\date{}

\maketitle

\begin{abstract}
Supervised machine learning algorithms have seen spectacular advances and surpassed human
level performance in a wide range of specific applications. However, using complex
ensemble or deep learning algorithms typically results in black box models, where the path
leading to individual predictions cannot be followed in detail. In order to address this
issue, we propose the novel "Cyclic Boosting" machine learning algorithm, which allows to
efficiently perform accurate regression and classification tasks while at the same time
allowing a detailed understanding of how each individual prediction was made.
\end{abstract}


{DOI:  10.1109/ICMLA.2019.00067}

\section{Introduction}

In many practical applications one is concerned with predicting a quantity $Y$ (target or
label) from a set of features $\bm{X}$, i.e. one needs to estimate the conditional
$p(Y|\bm{X})$ of the joint probability density distribution $p(Y,\bm{X})$ when the values
$\vec{x}$ of the feature variables $\bm{X}$ are observed. In supervised machine learning,
one uses e.g. neural networks like NeuroBayes \cite{NeuroBayes} to build a model for the
conditional, i.e. $\tilde{p}(Y|\bm{X},\vec{\theta})$, where $\vec{\theta}$ is a set of
parameters learned from  the training data and represents all observed knowledge of the
joint probability distribution $p(Y,\bm{X})$.

Although the predictions obtained from the machine learning model are in general very
accurate, the exact path how an individual prediction was calculated is typically not
observable in complex ensemble or deep learning models. Being able to explain how
individual predictions and decisions were made can be a mandatory legal requirement in
many sectors such as finance and insurance and is highly desirable in others such as
medicine or retail to build trust in the machine learning model.

In addition, most machine learning algorithms struggle to learn rare events, as they are
not representative of the bulk of the data and are often over-regularized, even though in
practical applications these effects may play a major role.

In order to address both the explainability and handling of rare effects, a novel machine
learning algorithm called "Cyclic Boosting" is proposed, which is able to learn a model
$\tilde{p}(Y|\bm{X},\vec{\theta})$ efficiently and accurately, while allowing to precisely
follow the path how individual predictions were made.

As will become apparent below, Cyclic Boosting can be categorized as a generalized
additive model \cite{GAM}, where the target $Y$ belongs to the family of exponential
distributions such as the Poisson, Gaussian, or Bernoulli distribution, together with a
suitable link function. As such, the Cyclic Boosting algorithm can be used in the
following three scenarios:
\begin{itemize}
\item{Multiplicative regression mode: $Y \in [0, \infty)$}
\item{Additive regression mode: $Y \in (-\infty, \infty)$}
\item{Classification mode: $Y \in [0, 1]$}
\end{itemize}

In the following, the multiplicative regression mode is described in detail first and for
the other two modes only modifications are highlighted afterwards.

\section{Literature Review}

Reaching human-level performance using machine learning techniques in specific
applications has also increased the interest in interpretability of algorithm-driven
decisions in recent years. However, due to the opaque nature of most of the complex models
it remains largely unexplained how these decisions are reached.

Several approaches exist to address this situation, e.g. Molnar \cite{molnar2019} gives a
good overview. In general, one can either use a complex black box model and subsequently
apply model-agnostic interpretation tools or build an explainable model.

For black box models, the importance of individual input features can be determined 
using e.g. Shapley's game-theoretic approach \cite{Shapeley1953, SHAP} or permutation
importance \cite{Breiman2001, eli5}. Google's {\em What-If} \cite{GoogleWhatIf} allows to
probe the behavior of a machine learning model if certain aspects or inputs are changed.
Partial dependency plots \cite{friedman2001} can be used to visualize the relationship
between the target and selected features and illustrate if e.g. the dependency is
non-linear or monotonous. Multiple visualization techniques have been developed in
particular to understand the way deep neural networks process images, such as partial
occlusion \cite{Zeiler2013} or saliency maps \cite{Simonyan2013}. These approaches are
undoubtedly invaluable tools to understand the inner workings of black box models as well
as the correlation between input features and predictions. However, they cannot address
the black box character on a fundamental level and therefore don't allow for fully
explainable models.

Surrogate models can be used to build explainable models out of black box models. For
instance, LIME \cite{lime} uses linear models to derive an explainable model which is
faithful locally around each prediction. While these predictions are locally explainable,
the model itself remains a black box model.

The simplest fully explainable model is the linear or logistic regression where the target
is represented by a sum of linear features and a Gaussian noise term
$\epsilon$: $y = \sum_i \alpha_i x_i + \epsilon$. Once all coefficients $\alpha_i$ are
determined from  data, each prediction can be evaluated in terms of the features and
their coefficients. However, this approach suffers from many short-comings, e.g. all
features are assumed to be linear with constant variance, as well as independent from each
other. Consequently, this simple approach is rarely sufficient in practical
applications.

General Linear Models (GLM) \cite{GLM} extend this approach and are useful for a wider
range of models. However, they still retain their linear character. More generally,
General Additive Models (GAM) \cite{GAM} replace the linear term with a function
$f_i(x_i)$ which allows for the modeling of non-linear effects. In general, GAMs are
described by $g(E[y]) = \beta_0 + \sum f_j(x_j)$, where $g$ is called the link function
and $f_j$ is some function which operates on the features $x_j$. In case of GLMs, $f$ is
constrained to be linear. While GAMs are not as easily interpretable as a simple linear
regression, they retain most of the benefits while allowing to model complex relationships
found in concrete application scenarios. Recently, Microsoft released {\em Interprete}
\cite{Interprete}, which includes a GAM with pairwise interactions for each feature
variable \cite{GAM_Microsoft}, i.e.
$g(E[y]) = \beta_0 + \sum_j f_j(x_j) + \sum_{i \ne j} f_{ij}(x_i,x_j)$.

Considering other supervised learning approaches, single decision trees \cite{Breiman2001}
are fully explainable, but are often not sufficiently performant in practical
applications. While amending them to ensemble methods by means of e.g. bagging or boosting
techniques can improve the performance of tree-based methods significantly, it greatly
reduces the explainability of the models as well. The same holds for support vector
machines (SVM) \cite{Boser1992}: For a linear kernel, the SVM weights define the
hyperplane which separates two classes. In case of low dimensional feature space, this can
be used to gain insights into the relative importance of the input variables. However, in
case of high-dimensional spaces this becomes more difficult and more general non-linear
kernels do not allow for easy interpretation of the SVM decision boundaries. Artificial
neural networks are typically considered as black box models due to their non-linear
transfer function modifying the output of individual neurons.

\section{The Cyclic Boosting algorithm}

\subsection{General approach}

The main idea behind Cyclic Boosting is that each individual feature $X_j$ from
$\bm{X} = (X_1, X_2, \ldots , X_p)$ contributes in a specific way to the prediction of the
target $\hat{Y}$. If all contributions can be calculated on a granular level, each
prediction $\hat{y_i}$ for a given observation $i$ can be transparently interpreted by
analyzing how much each feature $X_j$ for the observed values $x_{j,i}$ contributes to the
prediction.

To achieve the required granularity, each feature $X_j$ is first binned appropriately:
Categorical features retain their original categories, whereas continuous features are
discretized such that each bin has the same width (equidistant binning) or contains the
approximately same number of observations. In the following, bins are denoted by $b^k_j$,
i.e. bin $k = 1,..., n$ for feature $X_j$. During the training of the supervised machine
learning model, each feature, containing its various bins, is considered in turn and an
appropriate modification to the prediction $\hat{Y}$ of the target $Y$ is calculated. This
process is repeated iteratively until a stopping criterion is met, e.g. the maximum number
of iterations or no further improvement of an error metric such as the mean absolute
deviation (MAD) or mean squared error (MSE).

\subsection{Multiplicative regression mode}
\label{multiregmode}

In the multiplicative regression mode of Cyclic Boosting, the target variable is in the
range $Y \in [0,\infty)$. The predicted values of the target variable, denoted by
$\hat{y_i}$, are calculated from given observations $\vec{x}_i$ of a set of feature
variables $\bm{X}$ in the following way.

\begin{equation} \label{product}
\hat{y}_i = \mu \cdot \prod \limits_{j=1}^p f^k_j \quad \text{with}\; k=\{ x_{j,i} \in b^k_j\}
\end{equation}
Here, $f^k_j$ are the model parameters for each feature $j$ and bin $k$. For any concrete
observation $i$, the index $k$ of the bin is determined by the observation of $x_{j,i}$
and the subsequent look-up into which bin this observation falls. The global average $\mu$
is calculated from all observed target values $y$ taken across the entire training data.

If one assumes that the target variable $Y$ is generated as the mean of a Poisson (or more
general, negative binomial) distribution and the logarithm  $\ln$ is used as the link
function, eqn. \ref{product} can be inferred from the structure of a generalized additive
model by applying the inverse link function.

The model parameters $f^k_j$ are determined from the training data according to the
following meta-algorithm:

\begin{enumerate}
\item{Calculate the global average $\mu$ from all observed $y$ across all bins $k$ and
features $j$.}
\item{Initialize the factors $f^k_j \leftarrow 1$}
\item{Cyclically iterate through features $j = 1,...,p $ and calculate in turn for each
bin $k$ the partial factors $g$ and corresponding aggregated factors $f$, where indices
$t$ (current iteration) and $\tau$ (current or preceding iteration) refer to iterations of
full feature cycles as the training of the algorithm progresses:
\begin{equation} \label{factors}
g^k_{j,t} = \frac{\sum \limits_{x_{j,i} \in b^k_j} y_i}{\sum \limits_{x_{j,i} \in b^k_j} \hat{y}_{i,\tau}}
\;\; \mathrm{where} \; \; f^k_{j,t} = \prod \limits_{s=1}^t g^k_{j,s}
\end{equation}

\noindent
This means $g$ is the factor that is multiplied to $f_{t-1}$ in each iteration. Here,
$\hat{y}_\tau$ is calculated according to eqn. \ref{product} with the current values of
the aggregated factors $f$:
\begin{equation} \label{factors3}
\hat{y}_{i,\tau} = \mu \cdot \prod \limits_{j=1}^p f^k_{j,\tau}
\end{equation}
\noindent
To be precise, the determination of $g^k_{j,t}$ for a specific feature $j$ employs
$f^k_{j,t-1}$ in the calculation of $\hat{y}$. For the factors of all other features, the
newest available values are used, i.e., depending on the sequence of features in the
algorithm, either from the current ($\tau=t$) or the preceding iteration ($\tau=t-1$).
}
\item{Quit when stopping criteria are met at the end of a full feature cycle.}
\end{enumerate}

\noindent
This iterative cyclic optimization corresponds to a coordinate descent algorithm
\cite{Wright2015} with a boosting-like update of the factors $f$ and intrinsically
supports the modeling of hierarchical causal dependencies in the data by means of choosing
an appropriate feature sequence.

In order to increase robustness of the optimization and, if desired, reduce dependency on
the sequence of features, a learning rate $\eta$ can be added to the calculation of the
factors $f$ in eqn. \ref{factors} (with $\ln$ as link function):

\begin{equation}
\ln(\tilde g^k_{j,t}) = \eta_t \cdot \ln(g^k_{j,t}) \;\; \mathrm{where} \; \; \eta_t \in (0,1]
\end{equation}

\noindent
Here, $\eta$ is chosen as a small value at the beginning of the training ($t=1$) and is
then increased after each full feature cycle $t$ according to a linear or logistic
function until it reaches $\eta = 1$ for the maximal number of iterations, hence
$\tilde g^k_j \to g^k_j$ as the algorithm converges.

If the values $y$ follow a Poisson distribution, the Cyclic Boosting algorithm corresponds
to optimizing $\sum_i \frac{(y_i/\hat{y}_{i,\tau} - g^k_j)^2}{\sigma_i^2}$, i.e. $\chi^2$,
with $\sigma_i^2 = y_i/\hat{y}_{i,\tau}$ for all observations $i$ in each bin $k$ of
feature $j$. Since the bins of each feature variable are considered independently of each
other, the optimization is performed locally in each bin $b^k_j$. This has the benefit
that rare events can be learned effectively by the algorithm. While most machine learning
algorithms tend to over-regularize these effects, especially when they are far away from
the bulk of the respective distribution of observed feature variables $X_j$, choosing a
suitable binning allows to treat rare observations separately from the bulk of the
distribution of observed feature variables and hence allow accurate predictions even in
this case. However, the potentially low numbers of observations in such bins increase the
need for regularization methods in order to avoid learning wrong or spurious relationships
from data, i.e. reduce the risk of overfitting.

Although the cyclic consideration of all variables already accounts for correlations
between the different features, the learning of correlations between specific features can
be further improved by adding composed features with multi-dimensional binning, e.g.
built out of two or three of the original features. An example for this is described later
on in figure \ref{fig:store_td_2D}.

The binned feature-wise optimization of the Cyclic Boosting method also enables a natural
access to the introduction of sample weights. As an example, this can be used to put more
emphasis on the most recent past when predicting a target available as times series data.
Such a procedure can help to improve the forecast quality in case of trends or other
temporal changes in the data.

Owing to its straightforward structure based on fundamental arithmetic operations, Cyclic
Boosting can be trained efficiently on a large amount of data and parallelization of the
algorithm is possible without major obstacles.

\subsection{Regularization}
\label{conjugates}

The factors $f^k_j$ are iteratively updated according to eqn. \ref{factors}, where the
update rule has the form $g = \alpha / \beta$. As the Gamma distribution is the maximum
entropy probability distribution for a random variable $\xi$ for which
$E[\xi] = \alpha / \beta$ is fixed and greater than zero, the Gamma distribution is
assumed as a prior for the distribution of the factors $f^k_j$ in each bin $k$ of feature
$j$. Furthermore, the numerator and denominator of eqn. \ref{factors} have the form of the
maximum likelihood estimator for an i.i.d. random variable following a Poisson (or more
general negative binomial) distribution. These considerations motivate the description of
the individual contributions, i.e. the factors, to the prediction of a target variable
$Y \in [0,\infty)$ as conjugate distributions, the Gamma distribution being the conjugate
prior to the Poisson (or more general negative binomial) likelihood. Eqn. \ref{factors}
can hence be written as:

\begin{equation} \label{posteriors}
g^k_j = \frac{\alpha^k_j}{\beta^k_j}
\end{equation}
with
\begin{equation}
\alpha^k_j = \alpha_{\text{prior}} + \sum \limits_{x_{j,i} \in b^k_j} y_i \;\; \mathrm{and} \; \;
\beta^k_j = \beta_{\text{prior}} + \sum \limits_{x_{j,i} \in b^k_j} \hat{y}_i
\end{equation}

\noindent
The numerical values of the parameters of the prior Gamma distribution are chosen such
that the median of the Gamma distribution is 1, i.e. $\alpha_{\text{prior}} = 2$,
$\beta_{\text{prior}} = 1.67834$.

The definition of the factors in eqn. \ref{posteriors} exploits the fact that the mean of
the Gamma distribution can be expressed as $\alpha / \beta$. Instead, one could also
choose the median, which is generally a more robust point estimator and not as sensitive
to outliers as the mean.

\subsection{Smoothing}
\label{smoothing}

In most realistic applications, the observed data will be noisy and subject to statistical
fluctuations, assuming that missing, incomplete or wrong data have already been corrected
and common best practices for improving data quality have been observed. Regularizing the 
factors $f^k_j$ across bins $k$ for each feature $j$ will therefore improve the numerical
stability of the algorithm training. For categorical features, the factors in each
category can be regularized by determining appropriate Bayesian {\em a priori}
probabilities for each occurrence of the specific category of feature variable $X_j$. For
continuous features, smoothing functions such as splines or a suitable base of orthogonal
polynomials can be applied, which is equivalent to applying a low-pass filter to remove
high-frequency noise.

It should be noted that the range of the factors need to be transformed from $(0, \infty)$
to $(-\infty, \infty)$ before these smoothing approaches can be applied. This can be
achieved by taking the logarithm of the factors, i.e. $f^{\prime k}_j = \ln(f^k_j)$. In
order to be able to fit a smoothing function to the factors, the uncertainties
$\sigma_{f^{\prime k}_j}$ of each factor $f^\prime$ in each bin $k$ for feature $j$ can be
estimated from moment matching of the Gamma distribution to the log-normal distribution,
i.e. assuming that the uncertainties follow a Gaussian distribution after the logarithmic
transformation has been applied. This means the variance of the Gamma distribution is set
equal to the variance of the log-normal distribution:

\begin{equation}
\frac{\alpha}{\beta^2} = (e^{\sigma^2} - 1) \cdot e^{2(\mu + \frac{\sigma^2}{2})}
\end{equation}

\noindent
The mean of the log-normal distribution is then substituted by the mean of the Gamma
distribution: $e^{\mu + \frac{\sigma^2}{2}} = {\alpha}/{\beta}$.

\noindent
And finally, this leads to the following formula for the uncertainties:

\begin{equation}
\sigma^2_{f^{\prime k}_j} = \log (1 + \alpha^k_j) - \log (\alpha^k_j)
\end{equation}

After the smoothing of the factors has been performed, the factors are transformed back to
the original range (i.e. $(-\infty, \infty) \to (0, \infty)$) by applying the exponential
function as the inverse of the natural logarithm.

\subsection{Additive regression mode}

In the additive regression mode, with the range of the target variable being
$Y \in (-\infty, \infty)$, the formulae are modified such that:

\begin{equation} \label{summands}
\hat{y}_i = \mu + \sum \limits_{j=1}^p f^k_j  \quad \text{with}\; k=\{ x_{j,i} \in b^k_j\}
\end{equation}

\begin{equation}
f^k_{j,t} = \sum \limits_{s=1}^t g^k_{j,s} \;\; \mathrm{and} \;\; g^k_{j,t} = \sum \limits_{x_{j,i} \in b^k_j} y_i - \sum \limits_{x_{j,i} \in b^k_j} \hat{y}_{i,\tau}
\end{equation}

The conjugate distributions for the individual contributions to the prediction, in this
case the summands, follow a Gaussian function. Therefore, no transformation is needed
before smoothing.

\subsection{Classification mode}

In the case of (binary) classification, one aims to identify whether a given observation
$i$ belongs to a certain class or not. Hence the range of the target variable is in
$[0,1]$, which can be interpreted as the probability $p_i$ that this observation belongs
to the class ($p_i \to 1$) or doesn't belong to the class ($p_i \to 0$). In practical
applications, a suitable cut-off has to be defined which separates the two cases.

Noting that the odds, i.e. the ratio $\frac {p_i}{1-p_i}$, has the range $[0, \infty)$,
the same approach as the multiplicative regression mode can be used:

\begin{equation} \label{odds}
\frac{\hat{p}_i}{1 - \hat{p}_i} = \mu \cdot \prod \limits_{j=1}^p f^k_j  \quad \text{with}\; k=\{ x_{j,i} \in b^k_j\}
\end{equation}

Instead of a Gamma function, the conjugate prior for the factors is now a Beta function,
due to the binary nature of the setting, and the corresponding likelihood is a Bernoulli
distribution. Choosing  $\alpha_{\text{prior}} = 1.001$ and $\beta_{\text{prior}} = 1.001$
results in a uniform Beta distribution for the prior that drops sharply to zero at either
end of the interval $[0,1]$, which is helpful to avoid overconfidence with extreme
predictions. The parameters of the posterior Beta distribution are then calculated as:

\begin{equation} \label{alpha}
\alpha^k_j = \alpha_{\text{prior}} + \sum \limits_{x_{j,i} \in b^k_j} y_i  \;\; \mathrm{and} \;\;
\beta^k_j = \beta_{\text{prior}} + \sum \limits_{x_{j,i} \in b^k_j} 1 - y_i
\end{equation}

The factors and their uncertainties are in turn estimated from the mean (or median) and
variance of this Beta distribution, similar to the approach taken for the multiplicative
regression mode.

The performance of the algorithm can be improved by the inclusion of sample weights
according  to the following scheme:

\begin{equation}
w_i = 
\begin{cases}
1 - \hat{p}_i, & \text{if}\ y_i = 1 \\
\hat{p}_i, & \text{if}\ y_i = 0
\end{cases}
\end{equation}

Similar to the approach taken in boosting, i.e. the combination of several weak learners
into a strong one, this definition enforces the training process to put more emphasis on
observations that have been misclassified in the current state of the algorithm. Eqn.
\ref{alpha} then reads:

\begin{equation}
\alpha^k_j = \alpha_{\text{prior}} + \frac{\sum \limits_{x_{j,i} \in b^k_j} w_i \cdot y_i}{\sum \limits_{x_{j,i} \in b^k_j} w_i}
\end{equation}

\begin{equation}
\beta^k_j = \beta_{\text{prior}} + \frac{\sum \limits_{x_{j,i} \in b^k_j} w_i \cdot (1 - y_i)}{\sum \limits_{x_{j,i} \in b^k_j} w_i}
\end{equation}

Like in the multiplicative regression mode, the logarithm is then used to transform the
range $(0, \infty)$ to $(-\infty, \infty)$ and in turn the same approach to regularization
and smoothing can be taken.

\section{Example: Demand Forecasting}
\label{demandforecasting}

A very useful application of Cyclic Boosting's multiplicative regression mode is to
forecast future demand of individual products sold in a retail location. Hereby, demand is
influenced by promotions, price changes, rebates, coupons, and even cannibalization
effects within the assortment range. Furthermore, customer behavior is not uniform but
varies throughout the week and is influenced by seasonal effects and the local weather, as
well as many other contributing factors. Hence, even though demand generally follows a
negative binomial distribution \cite{Ehrenberg1959}, the exact values of the parameters
are specific to a single product to be sold on a specific day in a specific location or
sales channel and depend on the wide range of frequently changing influencing factors
mentioned above. 

Cyclic Boosting allows to efficiently calculate all relevant parameters to model the
demand of individual products, taking a wide range of influencing factors into account,
while at the same time allowing the operational business to track and understand how each
individual prediction was made.

\subsection{Data and algorithm training}

We use data from a Kaggle online competition \cite{kaggle_data} to demonstrate demand
forecasting with Cyclic Boosting and showcase its properties. The data set consists of the
fields date, store, item, and sales, the latter being the target to predict. There are
five years of historical data, from beginning of 2013 until end of 2017, for 10 different
stores and 50 different items. 

Besides store and item, we include several features describing trend and seasonality,
namely days since beginning of 2013 as linear trend as well as day of week, day of year,
month, and week of month. A list of all used features (one- and two-dimensional) can be
found in the legend of fig. \ref{fig:factors}. For example, two-dimensional features
including the variable "item" allow to learn characteristics of time series of individual
products.

\begin{figure}
\begin{center}
\includegraphics[scale=0.3]{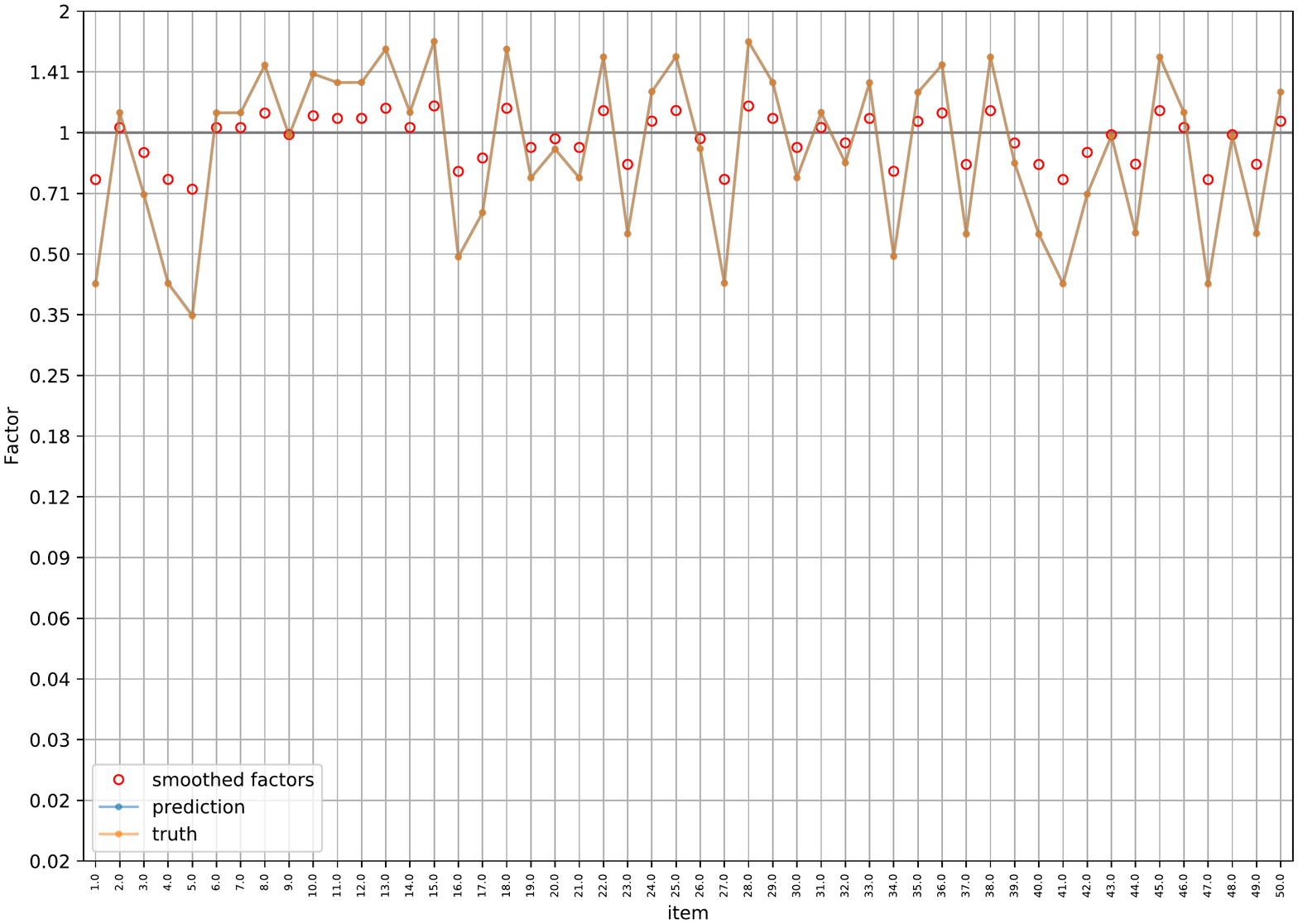}
\caption{\label{fig:item} Analysis of the feature variable "item" after the final
iteration. The shown data points indicate mean values of prediction $\hat{y}$ and ground
truth $y$ for each bin divided by the global mean (prediction hardly visible due to good
agreement). Note that the smoothed factors shown here are influenced by correlations to
all other features in the model as well.}
\end{center}
\end{figure}

Exemplarily, fig. \ref{fig:item} shows a detailed analysis of the factors for the feature
variable "item". Shown are the mean values of the prediction $\hat{y}$ after completion of
the training as well as the observed, true values $y$ in each bin divided by the global
mean. This visualization directly indicates possible deviations from the optimal fit
results in the different bins. Here, no significant deviations are present across the
whole range of values. Furthermore, the smoothed values of the factors, i.e. the actual
fitted parameters of the model, are shown. These differ from the mean values of the target
and prediction in the different bins divided by the global mean due to correlations with
other features.

\begin{figure}
\begin{center}
\includegraphics[scale=0.3]{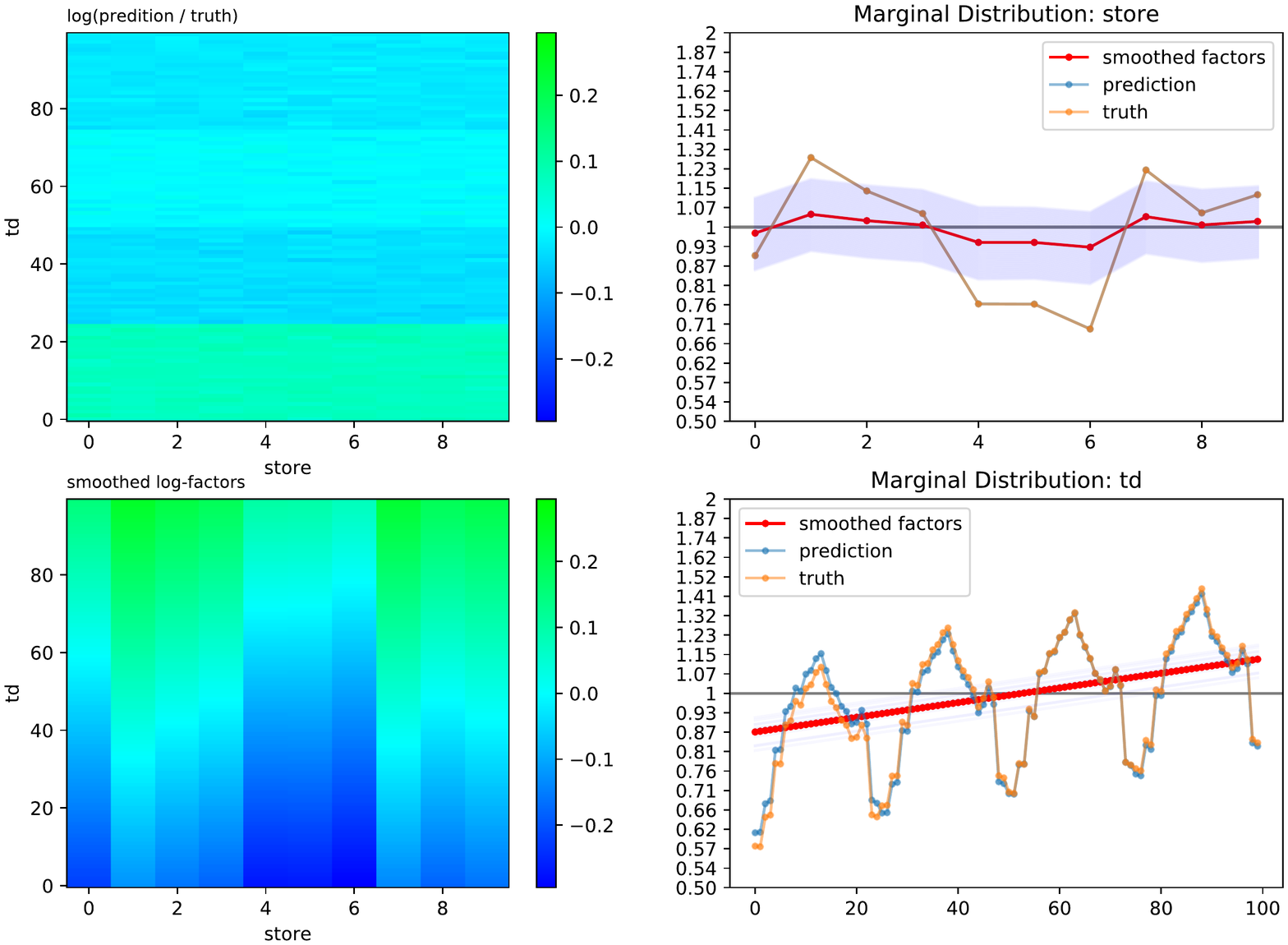}
\caption{\label{fig:store_td_2D} Analysis of the two-dimensional combination of the
features "store" and "td" after the final iteration. The upper left-hand plot shows the
two-dimensional deviations between prediction $\hat{y}$ and ground truth $y$, the lower
left-hand plot the smoothed factors, and the right-hand plots $\hat{y}$ and $y$ for the
two marginal distributions.}
\end{center}
\end{figure}

An example for a two-dimensional feature combination, namely  "store" and trend "td", is
shown in fig. \ref{fig:store_td_2D}. The upper left-hand plot shows a binned,
two-dimensional, color-coded visualization of the deviations between final predictions
and truth. The lower left-hand plot shows the smoothed values of the two-dimensional
factors, again visualized by means of color-coding. Here, one of the features is
categorical ("store") and the other one continuous ("td"), and the two-dimensional
smoothing is performed by means of grouping by the categorical feature dimension and
smoothing the continuous one. An alternative for a two-dimensional smoothing in case of
two continuous features consists in performing a truncated singular-value decomposition.
The two right-hand plots show the two corresponding marginal smoothed factor distributions
for the mean of the respective other dimension (solid red) and its individual categories
(transparent blue) as well as the marginal distributions for final predictions and
observed (true) values.

These examples show how Cyclic Boosting supports model development in terms of feature
engineering by means of analysis and individual preprocessing.

\subsection{Explanation of the predictions}

\begin{figure}
\begin{center}
\includegraphics[scale=0.55]{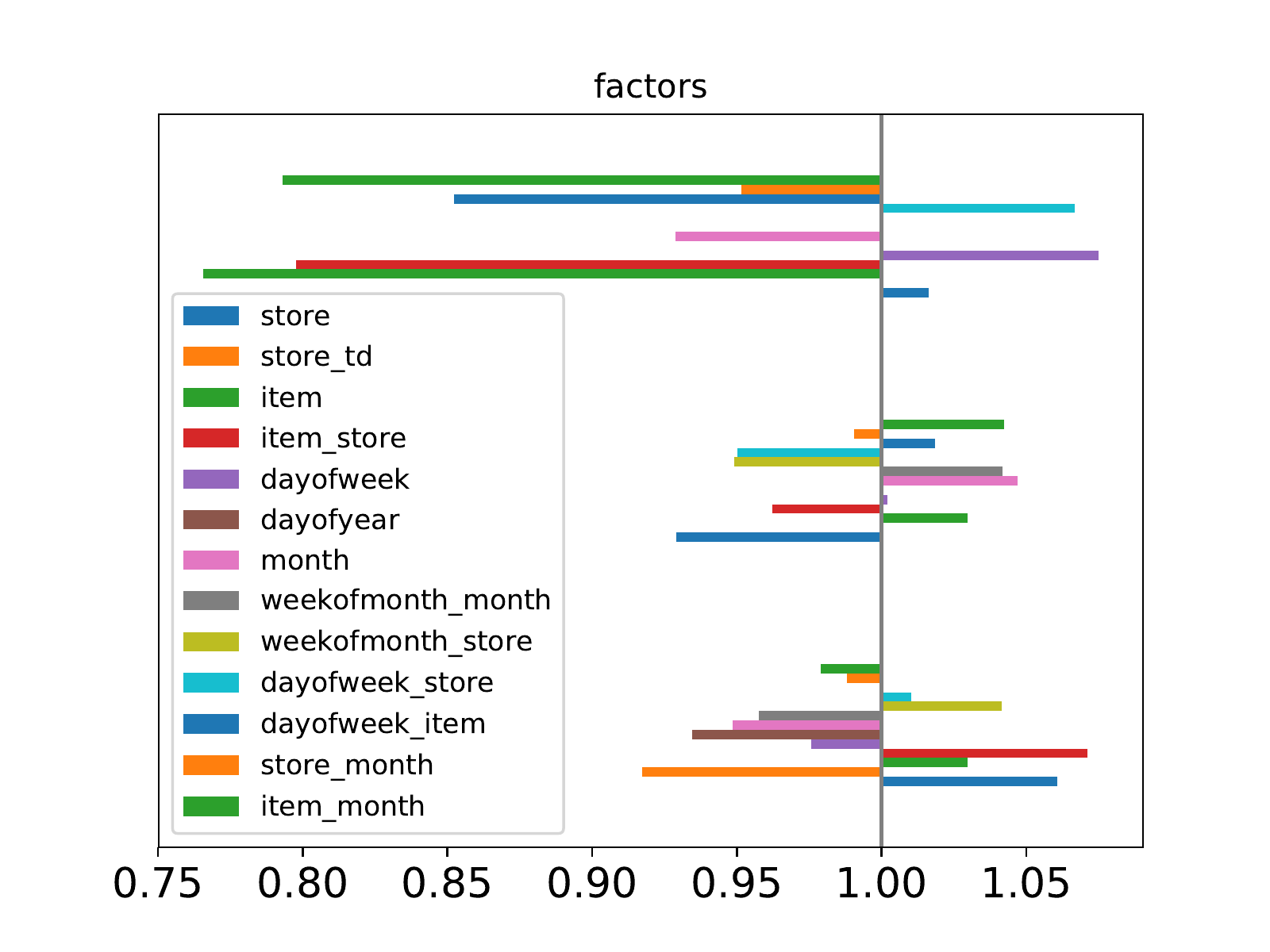}
\caption{\label{fig:factors} Illustration of the individual factors $f^k_j$ from which
the prediction $\hat{y_i}$ is calculated for three individual observations (displayed on
top of each other).}
\end{center}
\end{figure}

As stated above, one of the advantages of the Cyclic Boosting algorithm is that each
individual prediction can be interpreted and related to the feature variables used as
input, as shown in fig. \ref{fig:factors} for three different predictions $\hat{y_i}$. A
value of $f^k_j = 1$ implies that the importance of this particular feature is neutral
compared to the others, the strength of the deviation $f^k_j \neq 1$ indicates how
important a given feature is for the individual prediction. As the figure illustrates, the
importance of the individual features, from which the final prediction is calculated, can
vary significantly from one observation to the next.

\subsection{Results}

The published results of the competition correspond to a test period from beginning of
January to end of March 2018 \cite{kaggle_data}. However, the true observed values in this
test period are not publicly available. Since the main aim of this example is to
demonstrate that the Cyclic Boosting algorithm achieves at least comparable performance to
other machine learning approaches while retaining the benefit of fully explainable
predictions, the data until the end of 2016 were used for training the model and the first
three months of 2017 were taken as an independent test sample. This reproduces the
conditions of the competition as closely as possible and allows for a like-for-like
comparison of the published results and scores with the results obtained from this model.
Using the observed sales in the first three months of 2017 and comparing these to the
predicted values, the approach using Cyclic Boosting results in a symmetric mean absolute
percentage error of $\mathrm{SMAPE} \approx 13.20\%$. The same approach with a training
period until end of 2015 and prediction of the first three months of 2016 yields
$\mathrm{SMAPE} \approx 13.57\%$. The winning models of the competition have scores of
$\mathrm{SMAPE} \approx 13.84\%$ and $\mathrm{SMAPE} \approx 12.58\%$ for $34\%$ and
$66\%$ of the data set for the first three months in 2018. Given the upward trend of sales
and the larger training data set, we expect the $\mathrm{SMAPE}$ of our model in 2018 to
be at least comparable to the winning method. One can therefore conclude that Cyclic
Boosting can compete with other available algorithms in terms of forecast quality while
retaining full explainability of the individual predictions.

In order to demonstrate the robustness of Cyclic Boosting against variations of the number
of feature bins, which could be considered as hyperparameters, we both halved and doubled
the number of bins (from default value of 100) for the two utilized continuous features,
what resulted in changes in fourth place after the comma for $\mathrm{SMAPE}$ values.

As an additional remark, this simulated data set includes no information on prices,
promotions, or product hierarchy and also shows no dependency on events, like holidays,
weather, or other exogenous variables, for which the full potential of Cyclic Boosting
would come to fruition.

\section{Discussion: Causality}
\label{causality}

Causal dependencies play an instrumental role in any data generation process, and having reflected the underlying causal structure of the data, at least partially, in a supervised machine learning model is therefore beneficial for the generalization of the model and crucial for the interpretability of the learned correlations in terms of causal what-if scenarios for potential interventions \cite{PearlCausality,PearlML}.

\subsection{Individual causal effects}
\label{causalinf}

The estimation of individual causal effects, suffering from the problem of counterfactuals, needs a method capable of generalizations by means of sample attributes, which is one of the key strengths of machine learning. In the sense of potential outcomes \cite{Rubin}, a previously trained supervised machine learning model can be used to predict what-if scenarios for different values of one of its features.

Instead of the detour via two separate predictions, a direct prediction of absolute individual causal effects can be achieved with a machine learning algorithm capable of processing negative sample weights by subtracting one of the two potential outcome or what-if groups statistically from the other in the training. We call this technique statistical background subtraction. Due to its intrinsic feature binning, Cyclic Boosting in its additive regression mode (with adapted summand uncertainties) is ideally suited for this purpose: The external weights reflecting the respective membership to the two what-if groups ($+1$ and $-1$ in the case of certain memberships, positive and negative real values in the case of statistically estimated memberships \cite{Pivk_2005,PhysRevD.84.012003,PhysRevD.86.032007}), are simply applied to the sums over all samples in each feature bin in eqn. \ref{factors}, corresponding to the filling of the feature summand histograms. The accordingly weighted global average $\mu$ in eqn. \ref{summands} then corresponds to the average causal effect.

As an example, we look at the prediction of the individual causal effects of personalized coupons on customer demand (in terms of revenue) from the perspective of the retailer, and consider the simple case where unconfounded data from past random coupon assignments are available for the model training (see sec. \ref{confounding} for a discussion of confounding). Each training sample then represents an individual customer, the values $+1$ and $-1$ are used as sample weights for customers that did or did not receive a coupon, respectively, and the target is the corresponding revenue from that customer in some defined time period, e.g., a week. The resulting predictions of this model then correspond to the absolute individual causal effects of coupon sending on revenue. Note that this effect can be both positive or negative, an example for the latter being customers that would have bought also without coupon for a higher price.

Another benefit of the proposed combination of machine learning with statistical background subtraction, namely the focus of the model training on the causal effect to be learned, can be seen in our couponing example when considering that most of the customers just ignore a coupon completely, for instance by not showing any demand in the time period at hand, no matter if they received a coupon for it or not. Due to the random coupon sending, these customers are present to the same extent in both groups (with weights $+1$ and $-1$, respectively) of the data set used for the training, and are thus, thanks to the statistical background subtraction, effectively ignored in the model. In turn, the model can focus on learning the causal effect of the intervention on the target, rather than mainly learning the general dependencies of the target.

\subsection{Confounding}
\label{confounding}

There is one issue though with this approach of predicting individual causal effects with machine learning: Since the dependencies exploited by machine learning methods are merely statistical, the underlying causal structures of the data are not necessarily reflected in the learned model, because confounding effects can lead to spurious correlations overlaying the causal dependencies both between the different features and between these and the target.

The safest and most direct way to get rid of any confounding affecting an examined causal effect is via random assignment of the variable representing the cause, also known as randomized controlled trials \cite{RCT}. In case random assignment is not possible in a study or you are left with pure observational rather than interventional data, there is the need for a statistical method to avoid confounding of the data used for the training of the machine learning model, one example being independence weighting by means of inverse propensity scores \cite{propensity}. For this, each input sample for the training is weighted by the inverse of the corresponding propensity score value, which can, for example, be calculated by a separate machine learning model trained on the observational values of the variable representing the cause and containing all potential confounders as features.

Rather than eliminating confounding in the data, Cyclic Boosting also allows to impose causal assumptions during the model training process itself, either via exploiting the order of features, as mentioned in sec. \ref{multiregmode}, or by utilizing feature-specific smoothing functions over the factors for the different bins, in order to restrict the learning of the dependency between the target and potentially confounding or confounded features to defined parametric forms, e.g., monotonous functions. For potential confounders, the idea is to enforce the model to describe specific causal effects by other features, namely the true causes. For example, yearly seasonality in a demand forecasting model could be smoothed by a sinusoidal curve, leaving the description of distinct peaking structures to other features like holidays or promotions. Group-by smoothing can be used for two-dimensional features consisting of a cofounder and an interventional variable to stratify the confounders. The restriction of interventional features to specific functional forms, for example an exponential price-demand elasticity in retail demand forecasting, can also help to extrapolate beyond the range of the observations in the training data.

Besides the use case of causal inference described in sec. \ref{causalinf}, such incorporation of causal assumptions into a supervised machine learning model is also beneficial in terms of general forecasting quality, because it can improve the generalizability of the learned model. For a discussion of temporal confounding in time series forecasting see \cite{wick2021demand}.

\section{Conclusion}

A new machine learning algorithm, called Cyclic Boosting, was presented, which can be
categorized as a generalized additive model with a cyclic coordinate descent optimization
featuring a boosting-like update of parameters.

Cyclic Boosting addresses the challenge of prediction explainability on a fundamental
level: Rather than relying on black box approaches, individual predictions $\hat{y}_i$
for single observations should not only be accurate but also explainable. Each prediction
calculated using the Cyclic Boosting algorithm can be explained in terms of the strength
of each feature variable contributing to the prediction.

Furthermore, Cyclic Boosting facilitates (multi-dimensional) feature engineering and
enables the modeling of hierarchical causal dependencies and the prediction of rare
effects in the data. Thereby, overfitting is effectively avoided by means of
regularization and smoothing extensions.

\section*{}
Published at ICMLA 2019.

\noindent
\copyright 2020 IEEE. Personal use of this material is permitted. Permission from IEEE must be obtained for all other uses, in any current or future media, including reprinting/republishing this material for advertising or promotional purposes, creating new collective works, for resale or redistribution to servers or lists, or reuse of any copyrighted component of this work in other works. 

\noindent
Compared to the ICMLA paper (v2), we added (in v3) the discussion about causality in sec. \ref{causality}.

\bibliography{cb}
\bibliographystyle{ieeetr}

\end{document}